\documentclass[conference]{IEEEtran}
\IEEEoverridecommandlockouts

\usepackage{amsmath,amsfonts,bm}

\def\eqref#1{equation~\ref{#1}}

\def\1{\bm{1}}

\def\vr{{\bm{r}}}

\def\mA{{\bm{A}}}

\def\mC{{\bm{C}}}

\def\mE{{\bm{E}}}

\def\mG{{\bm{G}}}

\def\mI{{\bm{I}}}

\def\mR{{\bm{R}}}

\def\mZ{{\bm{Z}}}

\DeclareMathAlphabet{\mathsfit}{\encodingdefault}{\sfdefault}{m}{sl}
\SetMathAlphabet{\mathsfit}{bold}{\encodingdefault}{\sfdefault}{bx}{n}

\def\sP{{\mathbb{P}}}

\def\sV{{\mathbb{V}}}

\newcommand{\R}{\mathbb{R}}

\usepackage{amsmath,amssymb,amsfonts}
\usepackage{algorithmic}
\usepackage{graphicx}
\usepackage{textcomp}
\usepackage{xcolor}
\usepackage{times} 
\usepackage{tikz}
\usepackage{graphicx}
\usepackage{multirow}
\usepackage{adjustbox}
\usepackage{xspace}
\usepackage[dvipsnames]{xcolor}
\usepackage{booktabs}
\usepackage{dblfloatfix}
\usepackage[numbers,sort&compress]{natbib}
\usepackage[hyphens]{url}
\usepackage[capitalize,noabbrev]{cleveref}
\begin{document}

\title{GRAF-VLA: Gradient-Weighted Relevance Attribution for Flow-Matching Vision-Language-Action Models\\
\thanks{A. Iosifidis acknowledges funding by the European Union under the Horizon Europe research and innovation programme (Grant Agreement No. 101298363, Project OPERA). Views and opinions expressed are however those of the authors only and do not necessarily reflect those of the European Union or European Health and Digital Executive Agency. Neither the European Union nor the granting authority can be held responsible for them.}
}

\author{
\IEEEauthorblockN{Vidhi Agrawal\IEEEauthorrefmark{1}, Kostas Stefanidis\IEEEauthorrefmark{1}, Alexandros Iosifidis\IEEEauthorrefmark{2}\IEEEauthorrefmark{1}}
\IEEEauthorblockA{\IEEEauthorrefmark{1}\textit{Data Science Research Centre, Tampere University} \\
\IEEEauthorrefmark{2}\textit{ELLIS Institute Finland}, Finland \\
\{vidhi.agrawal, konstantinos.stefanidis, alexandros.iosifidis\}@tuni.fi}
}

\maketitle

\begin{abstract}

Vision-language-action (VLA) models increasingly combine pretrained vision-language backbones with diffusion- or flow-matching-based action decoders, whose cross-attention is rectangular and repeatedly evaluated across denoising steps. However, existing post-hoc explainability methods do not explicitly account for this structure. We introduce GRAF-VLA, which aggregates rectangular cross-attention relevance across steps, heads, and layers, combining it with self-attention relevance and Grad$\times$Input state attribution to produce relevance maps over visual, language, and state inputs in a single backward pass. We evaluate the resulting explanations through causal interventions, using per-step action-distance measurements and adaptive closed-loop rollouts on Meta-World MT50 and LIBERO, with Evo-1 as the evaluation model. Comparisons with attention-based baselines (AGVP) and random-region controls demonstrate that GRAF-VLA identifies task-relevant input regions, consistently outperforming random controls and achieving competitive performance against AGVP across both environments. Its results further highlight the benefits of explicitly accounting for different structures of diffusion-based VLA policies. Overall, we contribute an attribution method tailored to diffusion-based VLA policies and a causal evaluation methodology for assessing their explanations.
\end{abstract}   


\section{Introduction}
\label{sec:intro}
Vision-Language-Action (VLA) models have rapidly become a promising paradigm for general-purpose robot manipulation. OpenVLA \cite{pmlr-v270-kim25c} and RT-2 \cite{pmlr-v229-zitkovich23a}  combine pretrained Vision-Language model backbones with action-generation modules capable of translating natural language instructions and visual observations into low-level robot control actions. A growing share of recent VLAs replace simple regression or autoregressive action heads with diffusion or flow-matching decoders \citep{chi2025diffusionpolicy, black2025pi_0, liu2025rdt,lin2025evo}. These decoders iteratively refine a noised action sequence over many denoising steps conditioned on a vision-language context. This design consistently improves multi-modal action distribution modeling over single-shot decoders.

This performance gain comes at the cost of interpretability. VLA policies are deployed in physically consequential, safety-critical settings. Understanding why a model produces a given action, i.e., which visual regions, instruction words, or proprioceptive signals it relied on, is essential for detecting failure modes, building operator trust, and diagnosing correlations before deployment. Consider a robotic model handling fragile objects like glassware, or containers of a reactive substance being cleared for deployment after achieving a high success rate in training. A success-rate metric alone cannot reveal whether the model is grounding its grasp in the object's actual position and shape, or exploiting a co-occurring cue such as a support fixture or lighting condition that happened to correlate with object placement. The model's next failure may manifest as a dropped or crushed object rather than a benign miss, a failure that a success-rate curve cannot anticipate but an attribution method can.

Post-hoc explainability for vanilla Transformers is described in \cref{sec:related-transformers}, but no existing method is designed to explain a diffusion-based VLA's action decoder. Such methods assume a single square self-attention stack evaluated once. A diffusion-based action decoder is still built from Transformer blocks, but violates both assumptions. Its cross-attention is \emph{rectangular}: action-token queries belong to a token sequence that is structurally distinct and differently sized from the context sequence they attend to, thus relevance cannot be propagated as it is through self-attention where queries and keys are the same set. Moreover, it is evaluated \emph{repeatedly} over the denoising trajectory, with different refinement steps potentially contributing different semantic information to the final prediction \citep{park2024explaining}. Explainability work built specifically for Diffusion Models, in turn, has so far targeted text-to-image and image-to-image generators \citep{ICLR2025_c4a59e98, park2024explaining}, explaining generated image regions rather than a downstream action sequence conditioned jointly on vision, language, and proprioception. There is a lack of methods built for attention that is simultaneously rectangular, repeated, and directed at a non-visual output.

This paper aims to make diffusion-based VLA policies explainable by extending gradient-based Relevance Propagation to the rectangular, multi-step cross-attention structure of their action decoders. We propose \textbf{G}radient-weighted \textbf{R}elevance \textbf{A}ttribution For \textbf{F}lowmatching \textbf{VLA} models (\textbf{GRAF-VLA}), a post-hoc explainability method for VLA models whose action decoder is diffusion- or flow-matching-based. Our contributions are summarized as follows:

\begin{enumerate}
     \item We extend Relevance Propagation-style explainability to the rectangular cross-attention of diffusion- and flow-matching action decoders, using step-, head-, and layer-level aggregation to obtain a single relevance map.
    \item We fuse this cross-attention relevance with a self-attention path and a state-attribution relevance, giving jointly interpretable relevance over visual, language, and proprioceptive modalities in one backward pass.
    \item We introduce a causal evaluation of these maps: we extend masking/insertion faithfulness evaluation to closed-loop robot control, with per-chunk region re-selection and a size-matched random control, and complement it with an average per-step action-distance measurement.
    \item On Meta-World MT50, GRAF-VLA is the most faithful method under both masking and insertion until the effect saturates at highest fraction. On LIBERO, it is far more faithful than random regions and surpasses AGVP when more than $50\%$ of the patches are kept.
\end{enumerate}

The remainder of this paper is organized as follows. Section \ref{sec:related} reviews related works. Section \ref{sec:methodology} describes the targeted model architecture and the proposed method. Section \ref{sec:experiments} presents the experimental evaluation, and Section \ref{sec:conclusion} concludes the paper.

\section{Related Work}
\label{sec:related}

Early VLA models such as RT-1 \citep{Brohan-RSS-23} and RT-2 \citep{pmlr-v229-zitkovich23a} demonstrated that transferring internet-scale vision-language pretraining into robot control substantially improves generalization to novel objects and instructions. Subsequent work scaled this recipe further, with OpenVLA \citep{pmlr-v270-kim25c} introducing an open, reproducible 7B-parameter VLA trained across a broad multi-embodiment dataset. Along this line of work, the action decoder has evolved from simple discretized-token regression toward increasingly expressive generative heads motivated by the observation that robot action distributions are frequently multimodal.

Diffusion Policy \citep{chi2025diffusionpolicy} was among the first to demonstrate that framing action prediction as an iterative denoising process can improve multimodal action modeling and closed-loop task success over conventional regression-based decoders. This paradigm has since been extended to VLA models. $\pi_0$ \citep{black2025pi_0} combines a pretrained VLM backbone with a flow-matching action expert. RDT-1B \citep{liu2025rdt} further scales diffusion-based action generation to bimanual manipulation using a diffusion Transformer pretrained on large-scale multi-robot data. More recently, Evo-1 \citep{Lin_2026_evo1} introduces a lightweight VLA architecture that couples an InternVL3 backbone with a flow-matching action head. Evo-1 reports state-of-the-art performance on multiple benchmarks while using substantially fewer parameters than many existing VLA models \citep{Lin_2026_evo1}. We therefore adopt Evo-1 as a strong and recent baseline for studying explainability in flow-matching based VLA policies. These approaches differ in their vision-language backbones, action-head architectures, model scales, and training data, but share the use of iterative generative action prediction that motivates our method.


\subsection{Post-Hoc Explainability for Transformers}
\label{sec:related-transformers}
Attention-based interpretability has a long history in Transformer-based models. Raw attention weights have been criticized as an unreliable proxy for feature importance \citep{pmlr-v235-lopardo24a}, motivating Attention Rollout \citep{abnar2020quantifying}, which combines attention matrices across layers (with a residual connection correction) into a unified map to trace relevance back to input tokens. Generic Attention-model Explainability (GAE) \citep{hila2021gae} augments attention rollout with gradient-based weighting of each layer’s attention before propagation. Attention-Aware Layer-wise Relevance Propagation (AttnLRP) \citep{pmlr-v235-achtibat24a} instead extends LRP with Transformer-specific conservation rules for softmax, bilinear Q-K, A-V products, and LayerNorm, while backward-redistributing the output logit while preserving total relevance at every layer. Most recently, GMAR \citep{jo2025gmar} and GLIMPSE \citep{shen2025glimpse} further refine gradient-weighted attention aggregation for large Vision-Language Models. The former via a simpler scalar per-head gradient-magnitude weighting applied to raw attention, the latter via head and layer importance weighting. All are formulated for a square self-attention stack evaluated once, i.e., an encoder or an autoregressive decoder generating output in one pass, and none addresses rectangular or repeated passes over attention. We build directly on GLIMPSE's \citep{shen2025glimpse} gradient-weighted relevance-propagation machinery, extending it along both axes.

\subsection{Explainability for Diffusion Models}
A separate line of work targets explainability specifically for diffusion-based generative models. DAAM \citep{tang2023daam} was the first to aggregate raw cross-attention between text tokens and image patches across the denoising U-Net's spatial and temporal dimensions to produce word-level heatmaps, using no gradient information. $I^{2}$AM \citep{ICLR2025_c4a59e98} extends this to image-to-image latent diffusion via bidirectional cross-attention aggregation across time steps, heads, and layers. Closer to our approach, DF-CAM \citep{park2024explaining} adapts Grad-CAM to visualize which visual concepts a text-to-image model attends to at each denoising step, paired with DF-RISE, a perturbation-based gradient-free counterpart. All three models explain image generation. Their attention relates text or reference tokens to spatial image representations, not to a downstream action prediction attributed jointly to visual, textual, and robot state inputs through cross-attention that is both rectangular and repeated along the trajectory.

Within the VLA literature, Action-Guided Visual Pruning (AGVP) \citep{luo2026look} identifies task-relevant visual tokens by averaging raw action-to-vision attention  weights across action-token queries in shallow layers, pruning irrelevant tokens before deeper fusion. This design goal is orthogonal to ours and targets a non-diffusion action decoder. It is methodologically the closest prior attention-aggregation scheme to our relevance computation. We apply this attention-averaging strategy to the adopted backbone architecture, as a baseline.
\section{Methodology}
\label{sec:methodology}

\label{sec:prelim}

\subsection{Preliminaries}
GRAF-VLA targets VLA models that combine a vision-language model with a diffusion or flow-matching based action decoder. In this class of models, the vision-language backbone (typically a vision encoder followed by an $L_q$-layer self-attention language model) processes multiple camera views and a natural language instruction into a fused context sequence $\mC \in \R^{N \times d}$. A robot state vector is commonly appended to this sequence as well, extending it to $\mC \in \R^{M \times d}$, $M = N+1$. An $L_{\text{fm}}$-block action decoder then produces a horizon of $H$ future actions from an initial noise vector $\mZ_{q,0}$, refined toward a denoised action sequence over $T$ discrete denoising steps:
\begin{equation}
\label{fm}
    \mZ_{q,(t+1)} =  f_\theta(\mZ_{q,t}, \mC, t), \qquad t = 0, \dots, T-1,
\end{equation}
where $f_\theta(\cdot)$ is realized by cross-attention blocks in which the $\mZ_q$ noised action tokens attend, as queries, to the $M$-token context as keys and values. We design an explainability method for this type of VLA models. 

\begin{figure*}[t]
\centering
\includegraphics[width=0.85\linewidth]{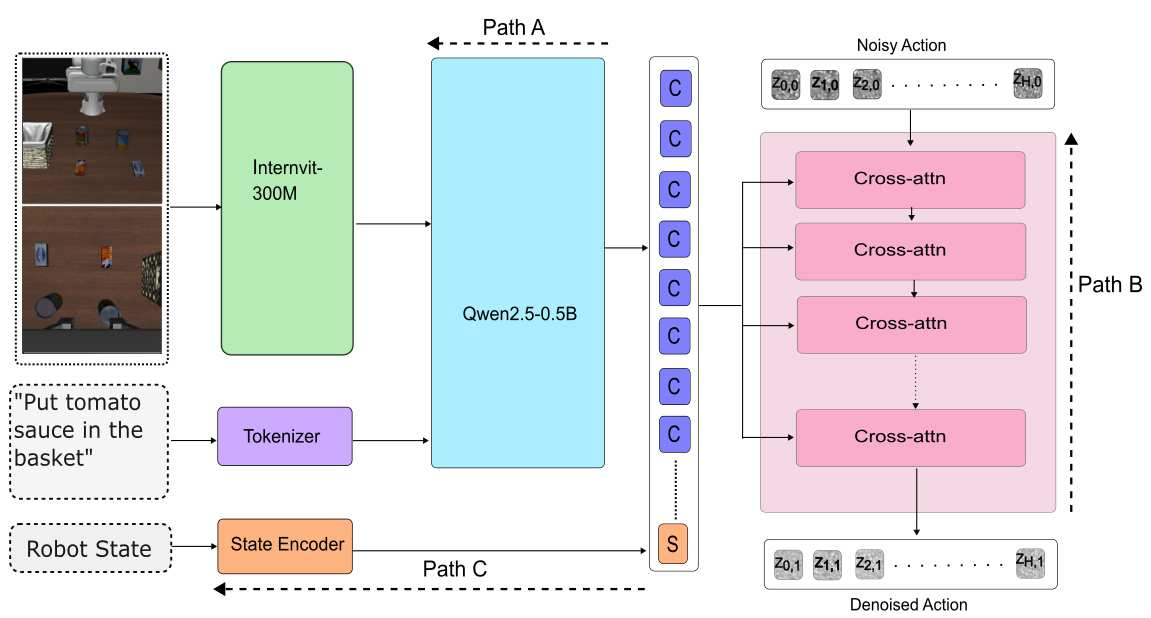}
\caption{VLA architecture highlighting Paths A, B and C used for relevance propagation by GRAF-VLA.}
\label{FIG:1}
\end{figure*}

For a model output $y$, \emph{relevance} measures an input representation's contribution to $y$ under the corresponding attribution procedure. Propagating relevance through the model assigns importance to intermediate representations, identifying information contributing to the produced action. We denote relevance propagation through backbone self-attention as \textbf{Path A}, through flow-matching decoder as \textbf{Path B}, and to the state input as \textbf{Path C} as shown in \cref{FIG:1}.

\subsection{Attribution target across diffusion steps}
We define $\phi$ as the sum of squared action-head outputs across all $T$ diffusion steps. Because $\mZ_{q,t}$'s relevance dependence on $\mZ_{q,(t-1)}$ is severed at every denoising step in \cref{fm} while $\mC$ remains live throughout, each step contributes an independent relevance branch rooted at $\mC$ and its own numerical action value. By the linearity of gradient accumulation across these branches, we obtain
\begin{align}
\phi &= \sum_{t=0}^{T-1}\Big\| f_\theta(\mZ_{q,t}, \mC, t) \Big\|_2^2, \\
\frac{\partial \phi}{\partial \mC} &= \sum_{t=0}^{T-1} \frac{\partial \big\lVert f_\theta(  \mZ_{q,t},\mC, t)\big\rVert_2^2}{\partial \mC}. 
\end{align}
The single backward pass on $\phi$ is mathematically equivalent to $T$ independent backward passes summed together, while requiring only one traversal of the $L_q$ self-attention layers upstream of $\mC$. All relevance computed below is thus attributed jointly across the full $T$ diffusion step trajectory from a single backward pass, rather than to one chosen step.

\subsection{Relevance Weighting}
\label{Relevance Weighting}
Although Path A and Path B differ in how relevance is propagated across layers, they share an identical process for converting captured attention scores and gradients into head- and layer-level relevance matrices. Both the $L_q$ layer language and $L_{\text{fm}}$ layer flow-matching submodels contain multi-headed attention. The attention head in each Transformer layer may not contribute equally to the model's output. Following Glimpse \citep{shen2025glimpse}, for a head $h$ in layer $l$, and with captured attention $\mA_l^h$ and its corresponding gradient $\mG_l^h$, we define per-head gradient-weighted relevance as
\begin{equation}
\bm{\Gamma_l}^h = \operatorname{ReLU}\big(\mG_l^h \odot \mA_l^h\big).
\end{equation}

Per-head importance is the ratio of relevance mass to total positive gradient mass, summed over all query-key index pairs $(i,j)$ that are softmax-normalized across heads at temperature $\tau_h$:
\begin{equation}
w_l^h = \operatorname{softmax}_h\left(\frac{\sum_{i,j}\bm{\Gamma}_l^h[i,j]}{\tau_h\cdot\sum_{i,j}\operatorname{ReLU}(\mG_l^h[i,j])}\right).
\end{equation}
Heads are fused into a single layer relevance matrix and then row-normalized to preserve probability mass:
\begin{equation}
\mE_l = \sum_h w_l^h\, \Gamma_l^h.
\end{equation}

To propagate relevance across layers, we use a combined weighting strategy that considers both gradient magnitude and layer depth. Layer importance combines total gradient magnitude with a depth prior that favors later layers with temperature $\lambda_d$:
\begin{align}
    g_l &= \Big\|\sum_{h} \mG_l^h\Big\|, \\
s_l &= \operatorname{softmax}_l(\lambda_d \cdot (l+1)), \qquad
\alpha_l = \frac{g_l\, s_l}{\sum_{l} g_{l}\, s_{l}}.
\end{align}
$\alpha_l$ balances empirical gradient evidence with architectural priors and allows strong gradient signals to override the depth bias when layers show exceptional importance for the prediction.

\subsection{Path A: Vision-Language's Self Attention Relevance}
Relevance is propagated additively across the $L_{\text{q}}$ self-attention layers. It is initiallized with Identity matrix ensuring that every token initially contributes only to itself:
\begin{align}
    \mR^{(0)} &= \mI_N, \\
\mR^{(l)} &= \mR^{(l-1)} + (\mI_N + \alpha_l \mE_l)\, \mR^{(l-1)}, l = 1, \dots, L_q.
\label{prop_qwen}
\end{align}
Here the identity term represents the network's residual connections. This lets accumulated relevance keep flowing forward while each layer's attention mixes in new relevance. Self-relevance becomes uninformative once propagation is complete, so rather than stripping it from the recursion itself, we apply it only to the last layer's matrix:
\begin{equation}
    \mR_q = \mR^{(L_q)} - \operatorname{diag}\big(\mR^{(L_q)}\big).
\end{equation}
Leaving the diagonal in would inflate every token's own score uniformly, diluting the softmax mass that should instead reflect influence from \emph{other} tokens.

The row-mean of $\mR_q$, i.e., $\vr_q = \operatorname{mean}_{i}\big(\mR_q[i,\,:]\big) \in \R^{N}$, then collapses this $N \times N$ relevance matrix to a single vector over the $N$ context positions giving a per-token relevance score aggregated across all $N$ query rows in the same $N$-dimensional context space that Path B's relevance vector $\vr_{fm}$ will be fused with.

\subsection{Path B: Flow Matching Cross Attention Relevance}
\label{Path B}
The $H$ action sequence queries attend to $\mC$ context tokens as keys/values. As Path A's square propagation in Eq. (\ref{prop_qwen}) is undefined for rectangular attention, and because $\mathbf{A}_l^{(t)}$ differs across the $T$ diffusion steps as the noisy action state evolves. Therefore, attention weights and gradients are captured independently at every one of the $T$ steps. Calculation of layer relevance $\mathbf{E}$ and $\alpha$ follows Sec. \ref{Relevance Weighting} for every denoising step separately. Layer fusion is performed independently at every denoising step and relevance is normalized over context tokens, before any averaging across steps:
\begin{align}
    \mR^{(t)} &= \frac{1}{H}\mathbf{1} + \sum_{l=1}^{L_f} \alpha_l^{(t)}\, \mE_l^{(t)}, \quad \mR^{(t)} = \frac{\mR^{(t)}}{\sum_j \mR^{(t)}_j}.\\
    \bar{\mR}_{\text{fm}} &= \frac{1}{T}\sum_{t=0}^{T-1} R^{(t)} \in \mR^{M \times H}.
\end{align}

The final column of $\bar{\mR}_{\text{fm}}$ corresponds to the auxiliary state token, which is attended to by the flow-matching head but never processed by Qwen. It is therefore separated from the visual/instruction context prior to fusion, i.e.,
\begin{equation}
    \mR_{\text{fm}} = \bar{\mR}_{\text{fm}}[\,1{:}N,:], \qquad \bar{\vr}_{\text{state}} = \bar{\mR}_{\text{fm}}[\,N{+}1,:],
\end{equation}
and used separately for state attribution (Sec. \ref{Path C}). The context component is then collapsed over its $H$ action-token queries by row-mean, i.e., $\vr_{fm} = \operatorname{mean}_{i}\big({\mR}_{\text{fm}}[i,\,:]\big) \in \R^{N}$, yielding the Path B relevance vector over the shared $N$-token context space.

\subsection{Relevance Fusion and Final Relevance Extraction}
\label{subsec:fusion}

The relevance vectors $\vr_q$ (Path A) and $\vr_{fm}$ (Path B) are independently standardized and linearly combined, i.e., 
\begin{equation}
    \begin{array}{l}
    \vr_{\text{final}} = \beta \cdot \frac{R_q - \mu(\vr_q)}{\sigma(\vr_q)} + (1-\beta) \cdot \frac{\vr_{fm} - \mu(\vr_{fm})}{\sigma(\vr_{fm})},
    \end{array}
\end{equation}
where $\mu(\cdot)$ and $\sigma(\cdot)$ are the scalar mean and standard deviation of a
vector over its $N$ context positions, applied by broadcasting to every element. Each path has a different raw scale due to factors such as attention entropy and gradient magnitude. Independent standardization of both vectors prevents either path from dominating.

The fused relevance vector $\vr_{\text{final}} \in \R^{N}$ spans visual patch, structural template (e.g. \texttt{Image,-,$i$,:}, \texttt{<img>}, \texttt{</img>}), and instruction tokens. To extract an interpretable relevance signal, we recover the index set $\sV$ of visual tokens directly from the vision-context token mask used in embedding fusion. The index set $\sP$ of input instruction tokens is recovered by locating the position following the final structural delimiter in the tokenized prompt and discarding the template tokens that precede it. Restricting to these two subsets and softmax-normalizing independently yields the final visual and prompt relevance distributions:
\begin{align}
    \textbf{\text{rel}}_{\text{visual}} &= \operatorname{softmax}\big(\vr_{\text{final}}[\mathbb{V}]\big), \\
    \textbf{\text{rel}}_{\text{prompt}} &= \operatorname{softmax}\big(\vr_{\text{final}}[\mathbb{P}]\big).
\end{align}
$\textbf{rel}_{\text{visual}}$ is reshaped into a $16 \times 16$ spatial grid per camera view, matching the resolution of the vision encoder's patch tokens after downsampling. They are bilinearly interpolated up to the original image resolution to produce a dense heatmap overlaid on the corresponding camera frame. $\text{rel}_{\text{prompt}}$ is reported per instruction token, giving a direct measure of which words in the language command the predicted action was most attributable to.

\subsection{Path C: Robot State Relevance}
\label{Path C}

The robot state input $\bm{s}$ is derived from simulation environment's observation vector, of which only the first $8$ entries are used by Evo1 \cite{Lin_2026_evo1}. They are exposed to the Feedforward network and encoded to $\mathbb{R}^{1 \times d}$ and attended to directly by the flow-matching action head. Because the state token is injected after Qwen's self-attention stack, Path A does not contribute to its attribution. We instead compute a Grad$\times$Input~\citep{shrikumar2017learning} attribution directly on $\mathbf{s}$. Applying the same $1/T$ step normalization used in Path A, the per-component contribution is
\begin{equation}
    c_k = \operatorname{ReLU}\!\Big(s \cdot \big(\tfrac{1}{T}\nabla_{s}\phi\big)\Big) \cdot m_k,
\end{equation}
which is then normalized into a distribution over the populated state components, i.e.,
\begin{equation}
    p_k = \frac{c_k}{\sum_{k'} c_{k'}}, \qquad k' \in \{1, \dots, 8\}.
\end{equation}
Here, $p_k$ determines how the state relevance is distributed among individual state components. However, it does not determine the total relevance assigned to the state input by the model relative to the visual and instruction inputs. We obtain this overall relevance from Path B's state component $\bar{\vr}_{\text{state}}$ (the auxiliary column separated out in Sec.~\ref{Path B}), which indicates how much of the flow-matching head's cross-attention relevance is attributed to the state token. We combine these two quantities multiplicatively to obtain the final relevance of each state component:
\begin{equation}
    \text{rel}_k^{\text{state}} = \operatorname{softmax}(\bar{\vr}_{\text{state}} \cdot p_k), \qquad k = 1, \dots, 8.
\end{equation}
As with the visual and prompt relevance of Sec.~\ref{subsec:fusion}, we separate two aspects of state attribution. $\bar{\vr}_{\text{state}}$ determines the overall relevance assigned to the state input at each timestep, while $p_k$ determines how this relevance is distributed among individual state components.

Paths A, B, and C attribute the produced action to all input modalities, i.e., visual patches and instruction tokens through the fused relevance of \cref{subsec:fusion}, and proprioceptive state through the formulation above. All are recovered from a single backward pass over the full denoising trajectory. 

\newcommand{\img}[1]{\includegraphics[width=0.18\textwidth]{images/task5_ep6_#1.png}}

\section{EXPERIMENTS}
\label{sec:experiments}

\subsection{Experimental Setup}
We evaluate our method on Evo-1 \citep{Lin_2026_evo1}, consisting of an InternVL3-1B \citep{zhu2025internvl3} vision-language backbone with a 14-layer Qwen2.5-0.5B \citep{bai2025qwen3} self-attention encoder and an 8-block flow-matching action decoder. Each camera view is patchified into a $32\times32$ grid and downsampled by pixel shuffle to 256 visual tokens. Together with the instruction, these form a context of $N=1024$ tokens, extended to $M=1025$ with the state token. The decoder predicts $H=50$ actions using $T_q=50$ query tokens over $T=32$ denoising steps, executing the first 15 before re-observation and re-planning. On Meta-World we set $\beta=0.9$, $\tau_h=0.9$, $\lambda_d^{\mathrm{A}}=0.9$, and $\lambda_d^{\mathrm{B}}=0.9$. On LIBERO we keep $\beta=0.9$ and $\tau_h=0.9$ but use $\lambda_d^{\mathrm{A}}=0.3$ and $\lambda_d^{\mathrm{B}}=1.3$, selected from a hyperparameter sweep across the grid explained in supplementary material Sec.1 .

We compare against AGVP \citep{luo2026look}, the closest existing method to ours on the application side. Methodologically, however, it is not directly compatible with our setting. It was proposed to prune visual tokens for a Qwen3-VL-based VLA with parallel action decoding, not to explain flow-matching action decoders. Its core idea, scoring visual tokens by raw attention from the action pathway averaged over layers and actions, does carry over to our backbone. We adapt it in two variants that differ only in the layers averaged. AGVP (shallow) uses the early layers (layers 2-8 of the 14 language-model layers and layers 1-4 of the flow-matching decoder), and AGVP (all layers) uses all layers of both. In both, the language model and flow-matching attention are fused with weight $\beta=0.5$. All baselines otherwise share GRAF-VLA's pipeline, from the saliency readout to the perturbation and initialization seeds, so differences come from how relevance is computed.

We evaluate on two simulated manipulation benchmarks. First, all 50 tasks of Meta-World MT50 \citep{pmlr-v100-yu20a}, spanning four difficulty tiers. Each task uses a scripted natural-language instruction and is evaluated for at most 400 environment steps, with object and goal configurations varied across episode resets. Second, the four task suites of LIBERO \citep{libero2023liu} (Spatial, Object, Goal and Long) which test spatial reasoning, object generalization, goal-conditioned behavior and long-horizon tasks, respectively. LIBERO observations contain two camera views (a third-person and a wrist camera), so relevance is computed jointly over the visual tokens of both views. Episodes start from the fixed LIBERO initial states, and the policy executes action chunks of 14 environment steps, up to 25 chunks per episode (95 for Long).

\subsection{Evaluation Metrics}
\label{sec:metrics}

We adopt the masking/insertion evaluation protocol of \citep{fong2017interpretable}, in which the causal importance of a saliency ranking is tested by selectively removing or retaining the regions it ranks highest and observing the effect on model behavior. For masking a lower success rate is better, while for inserting a higher success rate is better. We apply this protocol to visual input that the model attributes relevance to.

For the visual input, rather than modifying raw camera pixels as in the original formulation \cite{fong2017interpretable}, we mask or reveal patch embeddings directly inside the vision encoder, immediately after the patch-convolution projection and before positional encoding is added. This removes a selected patch's content at the earliest point in the model's computation reachable without altering the input image itself, while leaving the sequence structure (patch count, spatial layout) intact. A perturbation is specified by a fraction $k \in [0,1]$ of the $256$-token visual grid.


In the evaluation settings, all paired conditions (GRAF-VLA, random, AGVP baselines using shallow or all backbone layers, and the hybrid model) are initialized with the same noise vector, generated using a fixed random seed for each episode. This controls for diffusion-sampling stochasticity, allowing observed differences to be attributed to the perturbed content rather than variations in the initial noise.

\subsubsection{Case 1: Per-step action-distance}
\label{subsec:case1}
For a episode rollout, we advance the environment using the policy's clean, unperturbed input at every re-observation. At each re-observation, in addition to the action that drives the rollout, we separately query the policy on the same visual input that is perturbed at top-$k$ fraction, and record the $L_2$ distance between the two produced action sequences,
\begin{equation}
    d(k) = \big\lVert \hat{\mathbf{a}}_{\text{pert}}(k) - \hat{\mathbf{a}}_{\text{clean}} \big\rVert_2,
    \tag{28}
\end{equation}
without ever allowing ${\mZ}_{\text{pert}}(k)$ to affect the trajectory. This isolates \emph{prediction sensitivity}, i.e., how much perturbing the input changes what the model would have done, from any downstream, compounding effect of actually acting on a perturbed observation. $d(k)$ is averaged over every re-observation the episode takes and over $10$ independently-seeded episodes per task, per fraction $k \in \{0.0, 0.2, 0.4, 0.6,0.8, 1.0\}$.

\subsubsection{Case 2: Adaptive closed-loop rollouts}
Here, the predictions from perturbed inputs are executed. That is, at every re-observation, the top-$k$ (or random) region is re-identified from the current frame, the observation is perturbed accordingly, and the resulting action sequence (computed from the perturbed observation) is the one the robot actually carries out for the next $15$ environment steps before the cycle repeats. This measures whether continuously acting on a degraded observation compounds into task failure over a full closed-loop rollout. We report task success rate over $10$ episodes per task, per perturbation fraction $k \in \{0.1, 0.3, 0.5, 0.7, 0.9\}$, for all the GLIMPSE-guided, action guided pruning  and random-region conditions, using independent episode seeds but matched diffusion noise within each seed across conditions. 

\begin{figure*}[ht]
\centering

\begin{minipage}{0.84\textwidth}
\centering
\begin{tabular}{ccccc}
\multicolumn{4}{c}{\textbf{Overlay (visual relevance)}} \\[2pt]

\includegraphics[width=0.19\textwidth]{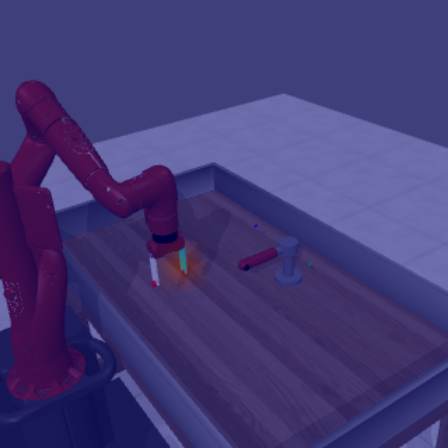} &
\includegraphics[width=0.19\textwidth]{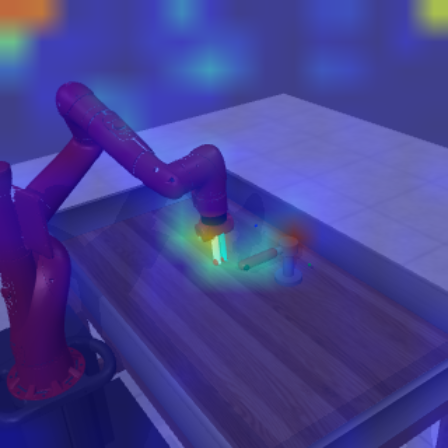} &
\includegraphics[width=0.19\textwidth]{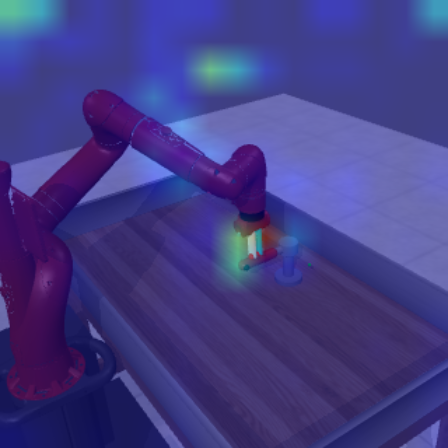} &
\includegraphics[width=0.19\textwidth]{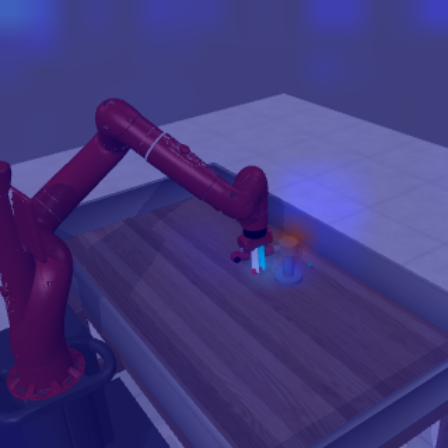} \\[4pt]

\multicolumn{4}{c}{\textbf{Prompt relevance}} \\[2pt]

\includegraphics[width=0.19\textwidth]{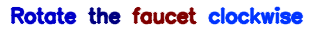} &
\includegraphics[width=0.19\textwidth]{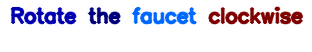} &
\includegraphics[width=0.19\textwidth]{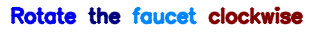} &
\includegraphics[width=0.19\textwidth]{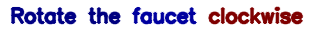} \\[4pt]

\multicolumn{4}{c}{\textbf{State relevance}} \\[2pt]

\includegraphics[width=0.19\textwidth]{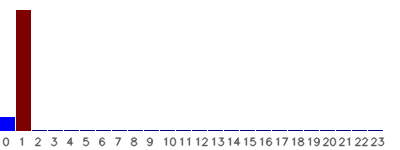} &
\includegraphics[width=0.19\textwidth]{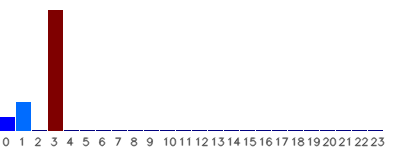} &
\includegraphics[width=0.19\textwidth]{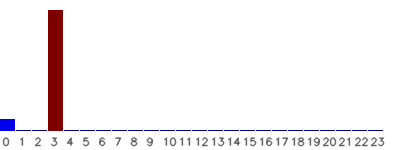} &
\includegraphics[width=0.19\textwidth]{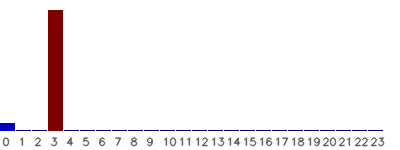} \\[2pt]

step 0 & step 15 & step 30 & step 45 
\end{tabular}
\end{minipage}%
\hspace{0.03\textwidth}%
\begin{minipage}{0.13\textwidth}
\centering
\begin{tikzpicture}
    \shade[bottom color=blue!90!black, top color=cyan]
        (0,0) rectangle (0.25,1.5);
    \shade[bottom color=cyan, top color=green!80!black]
        (0,1.5) rectangle (0.25,3);
    \shade[bottom color=green!80!black, top color=yellow]
        (0,3) rectangle (0.25,4.5);
    \shade[bottom color=yellow, top color=red!90!black]
        (0,4.5) rectangle (0.25,6);

    \draw[black, thin] (0,0) rectangle (0.25,6);

    \node[anchor=west, font=\scriptsize] at (0.3,6) {High};
    \node[anchor=west, font=\scriptsize] at (0.3,0) {Low};

    \node[rotate=90, anchor=south, font=\scriptsize] at (0.65,3) {Relevance};
\end{tikzpicture}
\end{minipage}
\caption{Relevance evolution over an episode of \texttt{faucet-close-v3} (``Rotate the faucet clockwise''), shown at re-observation steps 0, 15, 30 and 45. Top row: visual relevance overlaid on the camera frame; middle row: per-token prompt relevance; bottom row: per-input robot state relevance. The colorbar is shared across all three rows.}
\label{fig:saliency-evolution}
\end{figure*}

\begin{figure*}[ht]
\centering

\begin{minipage}{0.90\textwidth}
\centering
\setlength{\tabcolsep}{2pt}
\begin{tabular}{ccccc}
\multicolumn{5}{c}{\textbf{Overlay (visual relevance) -- agentview}} \\[2pt]
\img{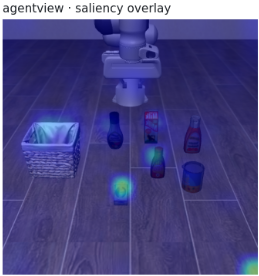} & \img{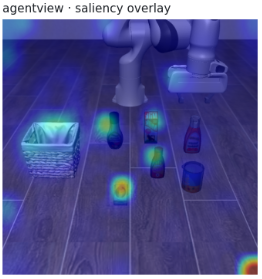} & \img{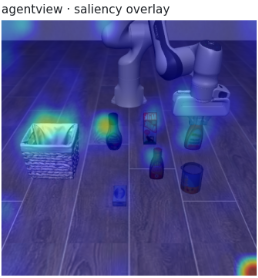} & \img{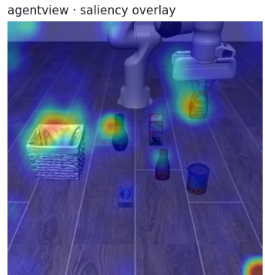} & \img{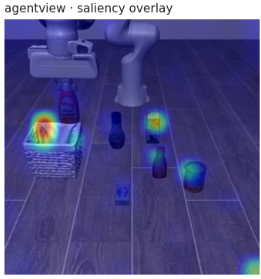} \\[4pt]

\multicolumn{5}{c}{\textbf{Overlay (visual relevance) -- wrist}} \\[2pt]
\img{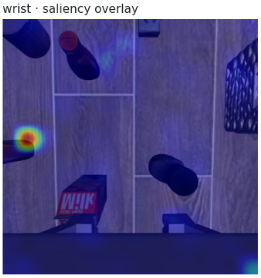} & \img{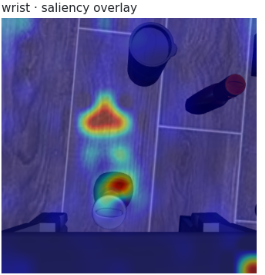} & \img{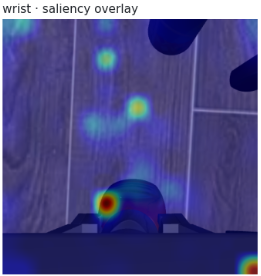} & \img{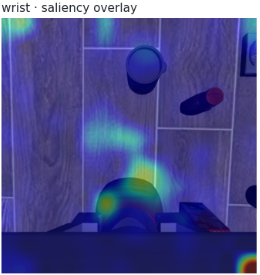} & \img{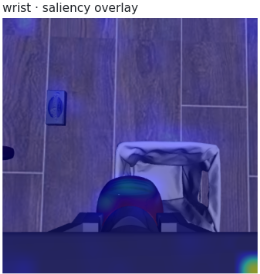} \\[4pt]

\multicolumn{5}{c}{\textbf{Prompt relevance}} \\[2pt]
\img{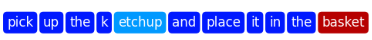} & \img{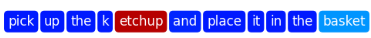} & \img{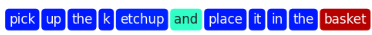} & \img{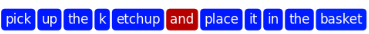} & \img{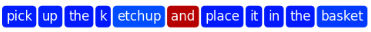} \\[4pt]

\multicolumn{5}{c}{\textbf{State relevance}} \\[2pt]
\img{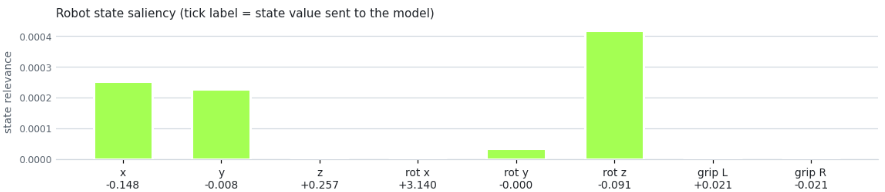} & \img{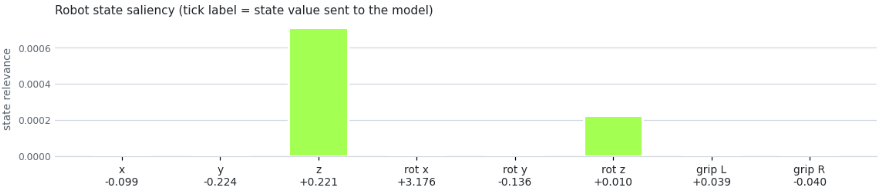} & \img{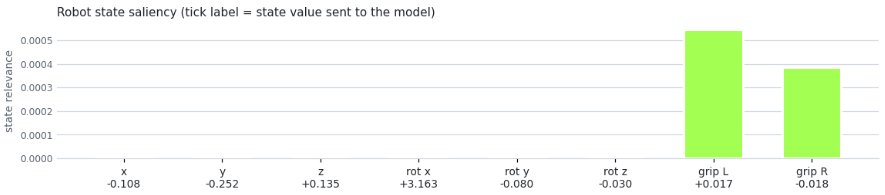} & \img{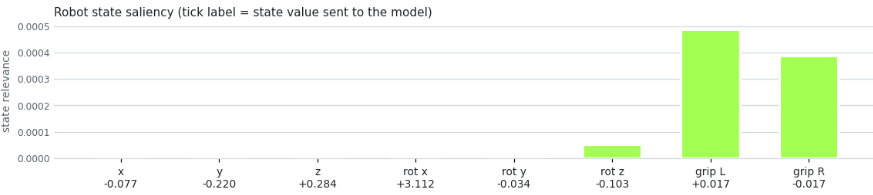} & \img{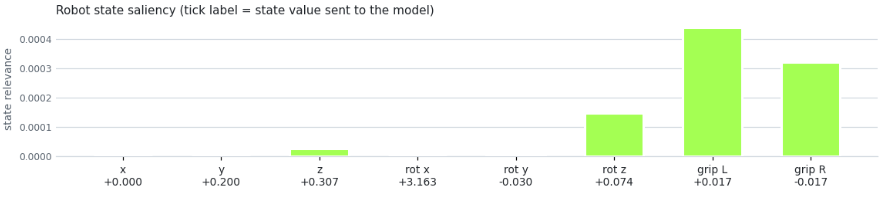} \\[2pt]

step 0 & step 30 & step 60 & step 90 & step 120
\end{tabular}
\end{minipage}%
\hspace{0.03\textwidth}%
\begin{minipage}{0.07\textwidth}
\centering
\begin{tikzpicture}
    \shade[bottom color=blue!90!black, top color=cyan]
        (0,0) rectangle (0.25,1.5);
    \shade[bottom color=cyan, top color=green!80!black]
        (0,1.5) rectangle (0.25,3);
    \shade[bottom color=green!80!black, top color=yellow]
        (0,3) rectangle (0.25,4.5);
    \shade[bottom color=yellow, top color=red!90!black]
        (0,4.5) rectangle (0.25,6);

    \draw[black, thin] (0,0) rectangle (0.25,6);

    \node[anchor=west, font=\scriptsize] at (0.3,6) {High};
    \node[anchor=west, font=\scriptsize] at (0.3,0) {Low};

    \node[rotate=90, anchor=south, font=\scriptsize] at (0.65,3) {Relevance};
\end{tikzpicture}
\end{minipage}
\caption{Relevance evolution over a LIBERO episode (``pick up the ketchup and place it in the basket''), shown at re-observation steps 0, 30, 60, 90 and 120. Rows 1--2: visual relevance overlaid on the agentview and wrist camera frames; row 3: per-token prompt relevance; row 4: per-input robot state relevance. The colorbar is shared across the rows.}
\label{fig:saliency-evolution-libero}
\end{figure*}


\begin{table}[b]
\centering
\small
\caption{Single-observation trajectory results: image perturbation on Meta-World, mean action distance $d(k)$ at each fraction $k$, across all 50 tasks. Mask: higher is better; insert: lower is better.}
\label{tab:so-image}
\begin{tabular}{l|cc|cc}
\hline
 & \multicolumn{2}{c|}{mask} & \multicolumn{2}{c}{insert} \\
\cline{2-3}\cline{4-5}
$k$ & GRAF-VLA & AGVP & GRAF-VLA & AGVP \\
\hline
0.0 & 0.00  & 0.00  & 27.57 & 27.57 \\
0.2 & 14.75 & 15.12 & 15.78 & 16.26 \\
0.4 & 18.05 & 19.03 & 9.82  & 9.68  \\
0.6 & 20.42 & 22.07 & 6.72  & 6.07  \\
0.8 & 23.69 & 25.29 & 4.33  & 3.45  \\
1.0 & 27.57 & 27.57 & 0.00  & 0.00  \\
\hline
\end{tabular}
\end{table}

\begin{table}[ht]
\centering
\caption{Image perturbation, adaptive rollout- metaworld, success rate averaged over 50 tasks.}
\label{tab:image-glimpse}
\begin{tabular}{p{0.7cm}p{0.6cm}|p{1.0cm}p{1.4cm}p{1.0cm}|p{0.9cm}}
\hline
Mode & per-cent & Random & AGVP (shallow) & AGVP (all layers) & GRAF-VLA  \\
\hline
mask & 10 & 65.8 & 35.4 & 43.6 & \textbf{18.8}  \\
mask & 30 & 49.2 & 17.0 & 30.6 & \textbf{6.6}  \\
mask & 50 & 33.2 & 10.2 & 16.0 & \textbf{3.2}  \\
mask & 70 & 13.8 & 9.0 & 8.4 & \textbf{5.0} \\
mask & 90 & 5.6 & 6.0 & 5.4 & \textbf{4.0} \\
\hline
insert & 10 & 7.0 & 11.8 & 6.6 & \textbf{22.2} \\
insert & 30 & 12.6 & 22.8 & 9.0 & \textbf{47.8} \\
insert & 50 & 29.6 & 41.4 & 24.8 & \textbf{56.6} \\
insert & 70 & 50.2 & 62.6 & 57.2 & \textbf{67.6} \\
insert & 90 & 63.4 & \textbf{69.9} & 69.8 & 68.6 \\
\hline
\end{tabular}
\end{table}

\begin{table}[ht]
\centering
\small
\caption{Image perturbation, adaptive rollout -- LIBERO, success rate averaged over the four suites (Spatial, Object, Goal, LIBERO-10).}
\label{tab:libero-glimpse-avg}
\begin{tabular}{p{0.7cm}p{0.6cm}|p{1.0cm}p{1.4cm}p{1.0cm}|p{0.9cm}}
\hline
Mode & per-cent & Random & AGVP (shallow) & AGVP (all layers) & GRAF-VLA  \\
\hline
mask & 10 & 84.5 & \textbf{41.5} & 54.2 & 59.5 \\
mask & 30 & 80.5 & \textbf{1.2} & 12.8 & 5.2 \\
mask & 50 & 52.5 & \textbf{1.0} & 1.8 & \textbf{1.0} \\
mask & 70 & 5.5 & 1.5 & 1.5 & \textbf{1.2}\\
mask & 90 & 1.2 & 1.2 & 0.8 & \textbf{0.5} \\
\hline
insert & 10 & 1.2 & \textbf{2.8} & 2.2 & \textbf{2.8} \\
insert & 30 & 4.8 & \textbf{56.0} & 17.0 & 37.0 \\
insert & 50 & 57.2 & \textbf{80.2} & 60.8 & \textbf{80.2} \\
insert & 70 & 80.0 & 80.8 & 78.2 & \textbf{86.0} \\
insert & 90 & 84.0 & 85.2 & 83.0 & \textbf{85.5} \\
\hline
\end{tabular}
\end{table}

\subsection{Results}
Figs.~\ref{fig:saliency-evolution} and~\ref{fig:saliency-evolution-libero} show the evolution of GRAF-VLA's relevance across an episode of Meta-World \texttt{faucet-close-v3} (``Rotate the faucet clockwise'') and a LIBERO-Object episode (``pick up the ketchup and place it in the basket''). In Meta-World, visual saliency is weak and confined near the faucet at step 0. It becomes strong on the gripper and the faucet handle by step 15 as the arm reaches the faucet, persists through step 30 while the gripper engages the handle, and weakens toward the arm at step 45 as the rotation completes. In LIBERO, agent-view saliency starts on the basket and the candidate objects, then follows the gripper and the object through the scene. Wrist-view saliency sharpens onto the target object as the arm approaches (step 30), stays on the object during grasping and transport (steps 60--90), and becomes weak and diffuse at step 120, when the basket fills the view and placement is under way.

Both visualizations highlight image corners that have no relation to the task. We attribute this to the known behavior of Vision Transformers, which store global information in a set of low-information patches whose local content is redundant with neighboring patches \cite{darcet2024vision}. These patches carry global context that the model relies on, and the gradient-based relevance correctly assigns them importance.

In both benchmarks, state saliency is concentrated on a few dimensions, a single dominant one throughout in Meta-World, and shifting toward the final dimensions in the later LIBERO steps. Prompt saliency is concentrated on one or two tokens that change with the subtask: ``faucet'' at step 0 and ``clockwise'' afterwards in Meta-World, and the object and goal tokens in LIBERO. The short, template-style instructions carry little discriminative information compared with the visual and proprioceptive inputs. Taken together, these trajectories show that GRAFVLA's saliency is not static across an episode but tracks the policy's shifting attention as the manipulation subtask changes.

Table \ref{tab:so-image} reports single-observation action distance \cref{subsec:case1} results on image input. Under masking, both methods produce a monotonically increasing distance as $k$ grows, correctly starting at $d(0){=}0$ and reaching the same fully-masked endpoint ($d(1.0)=27.57$) regardless of method. Under insertion, the reverse pattern holds. These shared endpoints, serve as an internal consistency check on the perturbation pipeline.

Table~\ref{tab:image-glimpse} shows that GRAF-VLA is the most faithful method on Meta-World under both masking and insertion, which indicates that its top-ranked regions are both necessary for the policy and sufficient to preserve its behavior. The advantage is largest at small perturbation budgets. The advantage vanishes at high fractions, where any method perturbs nearly all patches and the policy is either destroyed or recovered regardless of ranking. The comparison with the attention-only baselines is informative about depth: AGVP over all layers, which also uses deep layers but only raw attention, is clearly worse than AGVP over shallow layers. GRAF-VLA, whose depth prior favors the later layers, nevertheless outperforms both. This suggests that it is the gradient weighting, together with the flow-matching cross-attention path, that turns the deeper layers into spatially faithful relevance. Simple product or sum fusions of relevance matrices with AGVP do not help, so the attention-only map adds no complementary evidence.

On LIBERO (Table~\ref{tab:libero-glimpse-avg}), both GRAF-VLA and AGVP substantially outperform the random control, indicating that both identify task-relevant regions. AGVP performs slightly better when only a small fraction of patches is perturbed, whereas GRAF-VLA is comparable or superior at larger perturbation levels. The two-camera setup may enable the policy to compensate for patches removed from one view, reducing sensitivity to the exact ranking strategy. Further analysis of this effect is left to future work.

\section{CONCLUSIONS}
\label{sec:conclusion}

We presented GRAF-VLA, a gradient-based post-hoc explainability method for vision-language-action models with diffusion- or flow-matching-based action decoders. It extends relevance propagation to rectangular cross-attention repeated across denoising steps and adds attribution for proprioceptive state, producing saliency over all input modalities from a single backward pass. We evaluated visual explanations causally with single-step and closed-loop adaptive rollout experiments on Meta-World MT50 and LIBERO. On Meta-World, GRAF-VLA is more faithful than the attention-only baseline AGVP and a size-matched random control under masking and insertion. On LIBERO it is far above random and comparable to AGVP. Ablations show that attention over all layers is consistently worse than over shallow layers. Overall, GRAF-VLA extends post-hoc explainability beyond the single-pass, square-attention setting and provides a causal evaluation framework for iterative action decoders.
\section*{LLM Usage Statement}
In this work, we used generative AI tools (sonnet 5.5) for implementing methods, including debugging and correcting code in our evaluation and experiment scripts. We also used generative AI tools for drafting portions of the paper text (introduction, related works, experiments and conclusion). All AI-assisted code was reviewed, rechecked, and tested by the authors. AI-assisted paper text was reviewed and revised by the authors. We have reviewed all AI-assisted work for correctness and accuracy. We take responsibility for the final content of this work, including the text, claims, and artifacts produced with the aid of generative AI.
{
    \small
    \bibliographystyle{IEEEtran}
    \bibliography{main}
}

\clearpage
\twocolumn[
\begin{center}
{\LARGE \textbf{GRAF-VLA: Gradient-Weighted Relevance Attribution\\for Flow-Matching Vision-Language-Action Models}\\[1ex]
\large Supplementary Material\par}
\end{center}
]

\renewcommand{\thefigure}{S\arabic{figure}}
\renewcommand{\thetable}{S\Roman{table}}
\setcounter{section}{0}
\setcounter{figure}{0}
\setcounter{table}{0}

\section{Hyperparameter Selection}
\label{app:hparams}

\paragraph{Search protocol.} Both benchmarks were swept over the values in Table~\ref{tab:hp-grid}, and each configuration was scored by its closed-loop success rate under the adaptive-rollout masking/insertion protocol, with joint saliency over all cameras and the same fixed seeds (from 30-40) and diffusion noise for every configuration. On Meta-World, all configurations were evaluated on 50 tasks with 10 episodes each at 30\% masking and insertion. On LIBERO, all configurations were evaluated on 40 tasks with 10 episodes each at 50\% masking and insertion. We selected the configuration with the best success rates for insertion and worst ones for masking.
 Selection used the different environments, tasks and seeds as the final evaluation (seeds 42-52) on both benchmarks. The AGVP baseline uses its published settings ($\beta=0.5$) and was not tuned.
 
 \paragraph{Compute.} All experiments ran on NVIDIA GH200 GPUs, with the server (Evo-1 and relevance computation) on a 35\,GB GPU slice and the simulation client on an 18\,GB slice.
 
 \begin{table*}[!t]
\centering
\small
\caption{Hyperparameter values searched and selected.}
\label{tab:hp-grid}
\begin{tabular}{llccc}
\hline
Parameter & Flag & Searched values & Meta-World selected value & LIBERO delected value \\
\hline
$\lambda_d^{\mathrm{A}}$ & \texttt{depth\_temp}     & $-0.3, 0.3, 0.5, 0.7, 0.9, 1.3, 1.7$ & 0.9 & 0.3 \\
$\lambda_d^{\mathrm{B}}$ & \texttt{depth\_temp\_fm} & $-0.3, 0.3, 0.5, 0.7, 0.9, 1.3, 1.7$ & 0.9 & 1.3 \\
$\tau_h$                 & \texttt{head\_temp}      & $0.3, 0.5, 0.8, 0.9, 0.95$ & 0.9 & 0.9 \\
$\beta$                  & \texttt{qwen\_weight}    & $0.3, 0.5, 0.8, 0.9, 0.95, 1.0$ & 0.9 & 0.9 \\
\hline
\end{tabular}
\end{table*}

 \section{Single-observation trajectory for Libero}

Table~\ref{tab:libero-single-obs} reports the single-observation results on LIBERO, averaged over 200 observations (40 tasks, 5 seeds each). Both methods behave as a faithful attribution should: masking the top-$k$ most salient image tokens moves the action away from the unperturbed one, and inserting them restores it. GRAF-VLA already does this well. Masking only the top $60\%$ of its ranked tokens produces $96\%$ of the distance reached when all tokens are masked ($6.21$ of $6.49$), and inserting the top $80\%$ removes $90\%$ of the initial distance ($0.62$ of $6.49$). AGVP follows the same pattern and gives a further, modest improvement: it reaches slightly larger distances under masking at $k=0.2$, $0.4$ and $0.6$ (e.g.\ $5.87$ vs.\ $5.36$ at $k=0.4$) and slightly smaller distances under insertion at every $k$ (e.g.\ $0.49$ vs.\ $0.62$ at $k=0.8$), while the two methods are essentially tied under masking at $k=0.8$ ($6.64$ vs.\ $6.74$). In terms of the area under the $d(k)$ curve, the difference is $4.4\%$ for masking and $6.9\%$ for insertion, in favor of AGVP.

\begin{table*}[!t]
\centering
\small
\caption{Single-observation trajectory results: image perturbation on LIBERO, mean action distance $d(k)$ at each fraction $k$, across all four suites (40 tasks, 5 observations each, 5 seeds each).}
\label{tab:libero-single-obs}
\begin{tabular}{llcccccc}
\hline
Mode & Method & 0.0 & 0.2 & 0.4 & 0.6 & 0.8 & 1.0 \\
\hline
mask & GRAF-VLA & 0.00 & 3.50 & 5.36 & 6.21 & 6.74 & 6.49 \\
mask                   & AGVP     & 0.00 & 3.93 & 5.87 & 6.48 & 6.64 & 6.49 \\
\hline
insert & GRAF-VLA & 6.49 & 3.37 & 1.72 & 1.06 & 0.62 & 0.00 \\
insert                  & AGVP     & 6.49 & 3.02 & 1.60 & 0.98 & 0.49 & 0.00 \\
\hline
\end{tabular}
\end{table*}

\section{Additional Qualitative Results}
\begin{figure*}
    \centering
    \includegraphics[width=0.95\linewidth]{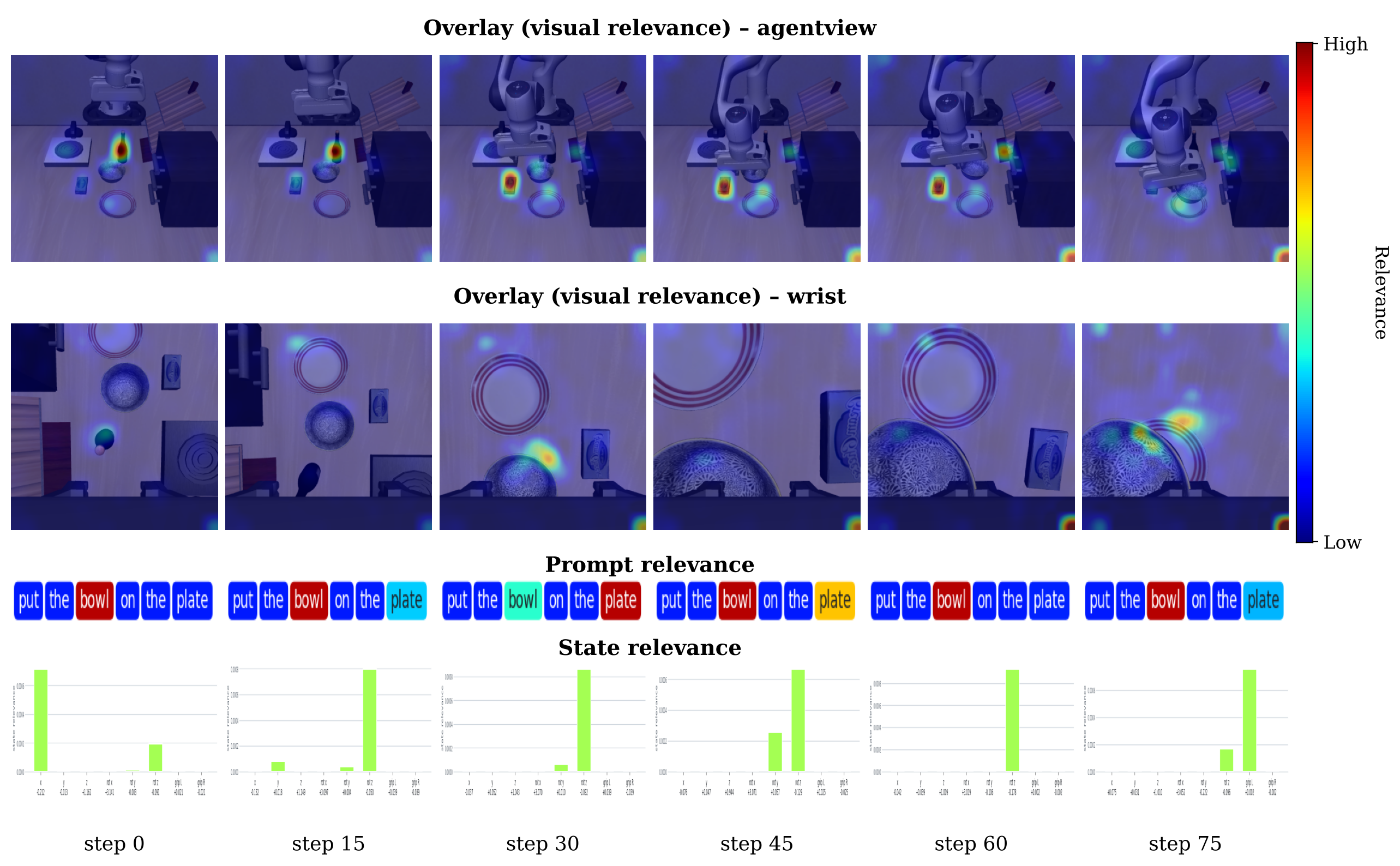}
    \caption{Relevance evolution over an episode of \texttt{Libero goal} (``put the bowl on the plate''), shown at re-observation steps 0-75.  Rows 1--2: visual relevance overlaid on the agentview and wrist camera frames; row 3: per-token prompt relevance; row 4: per-input robot state relevance. The colorbar is shared across the rows.}
    \label{fig:libero-goal-task9-ep9}
\end{figure*}

\cref{fig:libero-goal-task9-ep9} shows GRAF-VLA relevance over one LIBERO-Goal rollout for the instruction “put the bowl on the plate”. The highlighted regions change as the task progresses. At the start (steps 0–15), prompt relevance is concentrated on “bowl” and state relevance on the end-effector position, but the visual relevance sits mostly on the bottle in both cameras and only weakly on the bowl. This shows that the visual and language relevance are not tied to the same object. The bottle is a large, isolated object that may influence the planned motion, for example as an obstacle near the path of the arm. As the arm approaches the objects (steps 30–45), prompt relevance moves from “bowl” to “plate”, and visual relevance moves to the bowl rim and the objects near the gripper. At the end of the rollout (step 75), when the bowl is being placed, wrist relevance lies on the bowl–plate contact region and state relevance shifts to the gripper dimension. The visual maps also show relevance in the bottom-right corner of both cameras, a region unrelated to the task. This highlight is nearly absent in the first steps and becomes strong after step 30, when the gripper is close to the objects and the motion requires more careful control. This suggests that in this phase the model relies more on global scene information. We attribute the corner highlight to the known behavior of Vision Transformers, which store global information in a set of low-information patches whose local content is redundant with neighboring patches \cite{darcet2024vision}. Such a patch receives high attribution because the model reads global information from it, not because its pixels matter for the action. Its content is redundant with neighboring patches, so masking it in the deletion test or inserting it in the insertion test should have little effect on the policy.

\begin{figure*}
    \centering
    \includegraphics[width=0.95\linewidth]{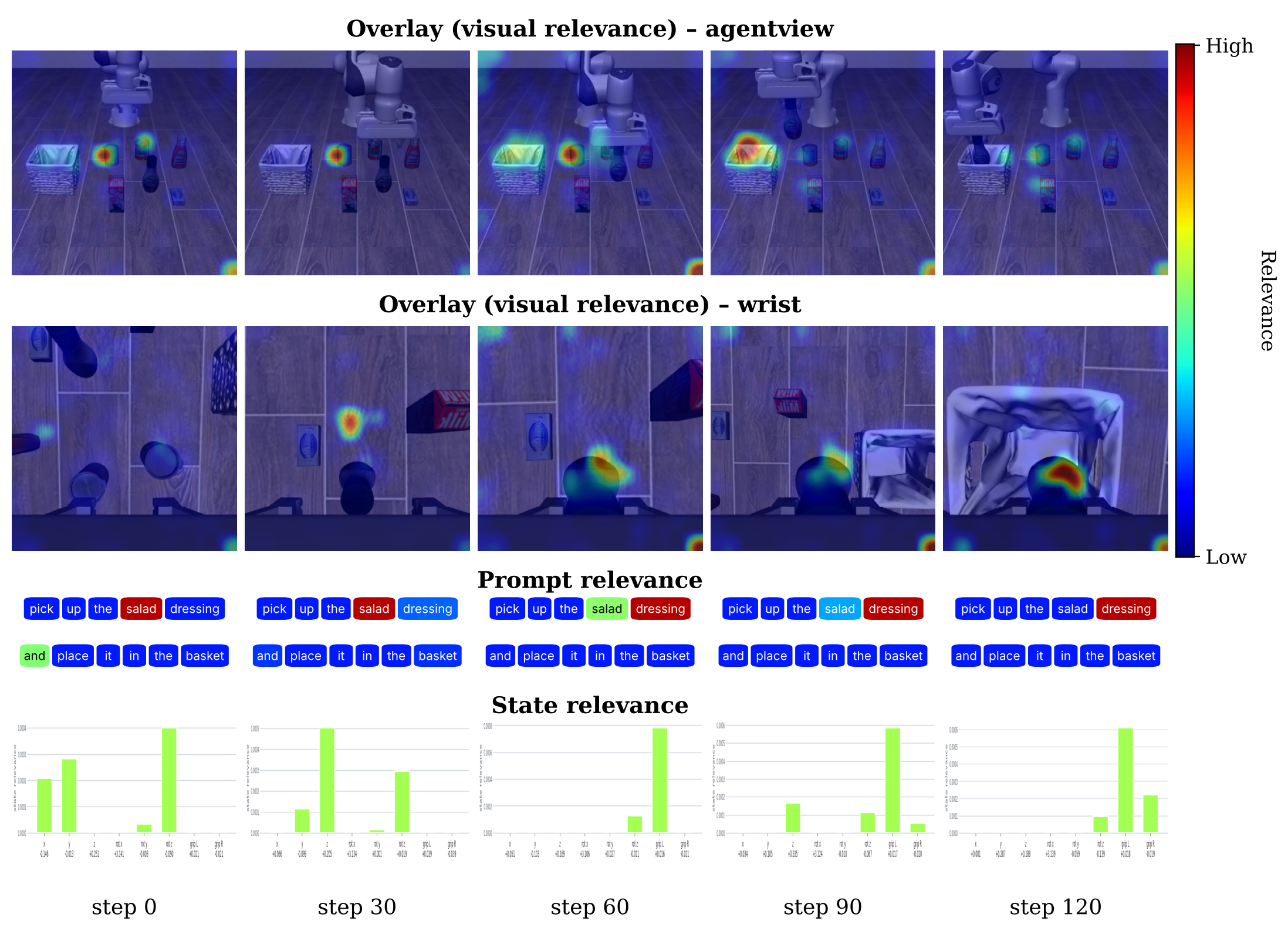}
    \caption{Relevance evolution over an episode of \texttt{Libero goal} (``pick up the salad dressing''), shown at re-observation steps 0-120.  Rows 1-2: visual relevance overlaid on the agentview and wrist camera frames; row 3: per-token prompt relevance; row 4: per-input robot state relevance. The colorbar is shared across the rows.}
    \label{fig:libero-obj}
\end{figure*}

\cref{fig:libero-obj} shows the same visualization for the instruction “pick up the salad dressing and place it in the basket” from Libero Object subtask category. Prompt relevance stays on the object phrase for the whole rollout, on “salad” at the start and on “dressing” from step 60 onward. It never moves to “place” or “basket”, so the basket information comes from the image. The visual relevance follows the sub-task. Before the grasp (steps 0–30), the agentview relevance is on objects near the gripper path, and the basket is only weakly highlighted. After the grasp (steps 60–90), the basket becomes one of the strongest regions, and the wrist view stays on the grasped bottle until the release (step 120). The state relevance also follows the phase of the motion. It is on the position and orientation dimensions while the arm moves over the objects (steps 0–30), and on the gripper dimensions from step 60, with both finger dimensions active around the release (steps 90–120). As in the LIBERO-Goal example, the relevance at the start is not always on the target. At steps 0 and 30 the agentview relevance is on a can next to the basket and not on the salad dressing bottle. The corner highlight also appears here, weak at the start and strong from step 60, which matches the explanation given for the LIBERO-Goal example.

\begin{figure*}
    \centering
    \includegraphics[width=0.95\linewidth]{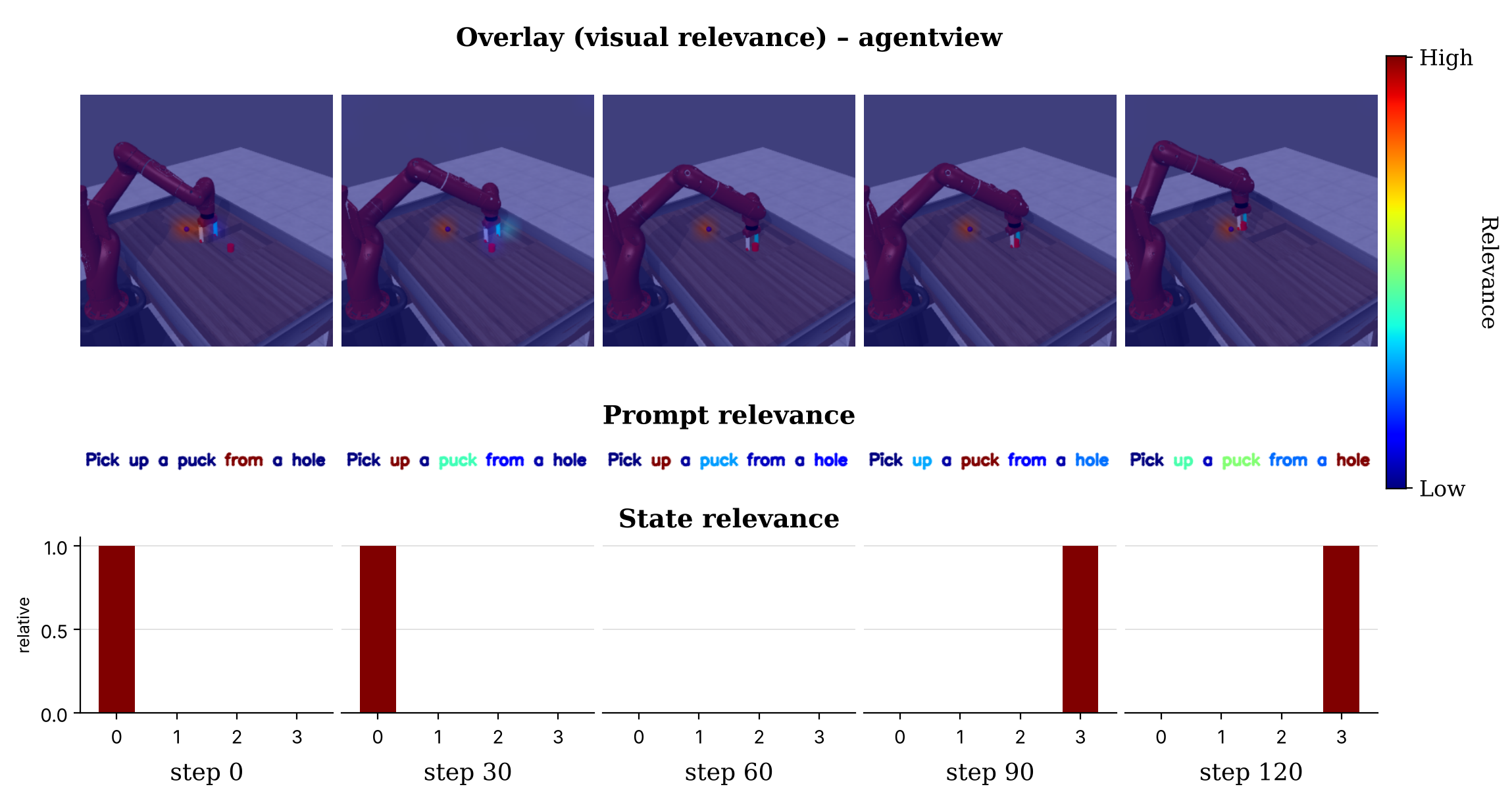}
    \caption{Relevance evolution over an episode of \texttt{pick-out-of-a-hole-v3} (``Pick up a puck from a hole''), shown at re-observation steps 0, 30, 60, 90 and 120. Top row: visual relevance overlaid on the camera frame; middle row: per-token prompt relevance; bottom row: per-input robot state relevance. The colorbar is shared across all three rows.}
    \label{fig:mataworld}
\end{figure*}
\cref{fig:mataworld} shows the same visualization for the Meta-World task “Pick up a puck from a hole” (pick-out-of-hole), with the agentview camera only. The relevance moves along the instruction as the episode proceeds. At step 0 the strongest word is “from”, at steps 30–60 it is “up”, at step 90 it is “puck”, and at step 120 it is “hole”. The visual relevance stays on the hole in every frame and is also on the gripper and the puck at steps 0–30, when the gripper is approaching the puck. The state relevance is on the first state dimension at steps 0–30 and on the fourth dimension at steps 90–120, and at step 60 it is close to zero for all four dimensions shown. Unlike the LIBERO examples, there is no relevance in the image corners in any frame. The Meta-World scene is simple, with one small object, one target and a fixed background. This suggests that the model does not need to store global information in extra patches here. In the LIBERO scenes, which have several objects and a cluttered table, the corner highlight appears and grows during the episode, and in this scene it never does. This agrees with our explanation of the corner highlight as a global-information patch, but we have not tested this directly.

\end{document}